\documentclass[11pt,a4paper]{article}

\usepackage[hyperref]{acl2017}
\usepackage{times}
\usepackage{latexsym}
\usepackage{url}
\usepackage{times}
\usepackage{helvet}
\usepackage{courier}
\usepackage[textsize=footnotesize]{todonotes}

\usepackage{url}
\usepackage{latexsym}
\usepackage{subcaption}
\usepackage{graphicx,verbatim}
\usepackage[rflt]{floatflt}
\usepackage{epsfig,times,graphicx}
\usepackage{amsmath,amssymb,amsopn,algorithm,algorithmic,theorem,float,bbm,bm,enumerate,color,multirow}
\usepackage{rotating}
\usepackage{array}

\newcommand{\beq}{\begin{equation}}
\newcommand{\eeq}{\end{equation}}

\newcommand\I{\mathbb{I}}

\newcommand\R{\mathbb{R}}

\newcommand{\g}{\mathbf{g}}

\renewcommand{\d}{\mathbf{d}}

\renewcommand{\u}{\mathbf{u}}

\newcommand{\w}{\mathbf{w}}

\renewcommand{\t}{\mathbf{t}}
\newcommand{\h}{\mathbf{h}}

\newcommand{\cT}{{\cal T}}

\newcommand{\cB}{{\cal B}}
\newcommand{\cD}{{\cal D}}

\newcommand{\cW}{{\cal W}}

\newcommand{\bD}{\mathbf{D}}

\newcommand{\bW}{\mathbf{W}}

\newcommand{\bT}{\mathbf{T}}
\newcommand{\bH}{\mathbf{H}}

\newcommand{\E}{\mathbb{E}}

\aclfinalcopy

\title{DocTag2Vec: An Embedding Based Multi-label Learning Approach for Document Tagging}

\author{Sheng Chen$^\sharp$\thanks{\ This work was done when the author was an intern at Yahoo.}, Akshay Soni$^\dag$, Aasish Pappu$^\ddag$, Yashar Mehdad$^\S$  \\
   $^\sharp$University of Minnesota-Twin Cities, Minneapolis, MN 55455, USA  \\
$^\dag$Yahoo Research, Sunnyvale, CA 94089, USA and $^\ddag$New York, NY 10036, USA \\
$^\S$Airbnb, San Francisco, CA 94103, USA \\
  {\tt chen2832@umn.edu \{akshaysoni,aasishkp\}@yahoo-inc.com}  \\
  {\tt yashar.mehdad@airbnb.com}
}

\date{}

\hypersetup{draft}
\begin{document}


\maketitle
\begin{abstract}
Tagging news articles or blog posts with relevant tags from a collection of predefined ones is coined as document tagging in this work. Accurate tagging of articles can benefit several downstream applications such as recommendation and search. In this work, we propose a novel yet simple approach called DocTag2Vec to accomplish this task. We substantially extend Word2Vec and Doc2Vec---two popular models for learning  distributed representation of words and documents. In DocTag2Vec, we simultaneously learn the representation of words, documents, and tags in a joint vector space during training, and employ the simple $k$-nearest neighbor search to predict tags for unseen documents. In contrast to previous multi-label learning methods, DocTag2Vec directly deals with raw text instead of provided feature vector, and in addition, enjoys advantages like the learning of tag representation, and the ability of handling newly created tags. To demonstrate the effectiveness of our approach, we conduct experiments on several datasets and show promising results against state-of-the-art methods.

\end{abstract}

\section{Introduction}
Every hour, several thousand blog posts are actively shared on social media; for example, blogging sites such as Tumblr\footnote[1]{tumblr.com} had more than 70 billion posts by January 2014 across different communities ~\cite{chang2014tumblr}. In order to reach right audience or community, authors often assign keywords or ``\#tags'' (hashtags) to these blog posts.  Besides being topic-markers, it was shown that hashtags also serve as group identities~\cite{bruns2011use}, and as brand labels~\cite{page2012linguistics}. On Tumblr, authors are allowed to create their own tags or choose existing tags to label their blog. Creating or choosing tags for maximum outreach can be a tricky task and authors may not be able to assign all the relevant tags. To alleviate this problem, algorithm-driven document tagging has emerged as a potential solution in recent times. Automatically tagging these blogs has several downstream applications, e.g., blog search, cluster similar blogs, show topics associated with trending tags, and personalization of blog posts. For better user engagement, the personalization algorithm could match user interests with the tags associated with a blog post. 

\begin{figure*}[hbt!]
	\centering
	\begin{subfigure}{0.40\textwidth}
		\centering
		\includegraphics[width=0.9\textwidth]{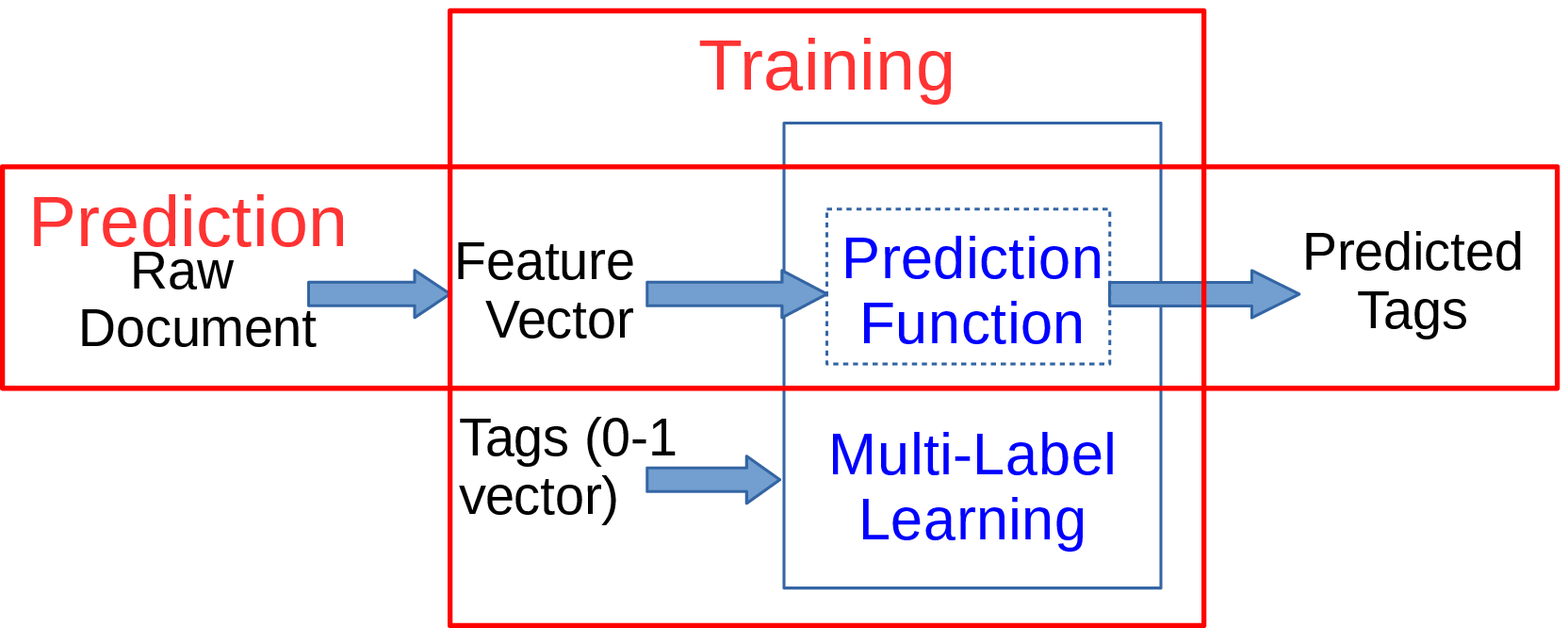}
        \caption{standard multi-label learning}
        \label{fig:sys1}
	\end{subfigure}
	\begin{subfigure}{0.55\textwidth}
		\centering
		\includegraphics[width=0.9\textwidth]{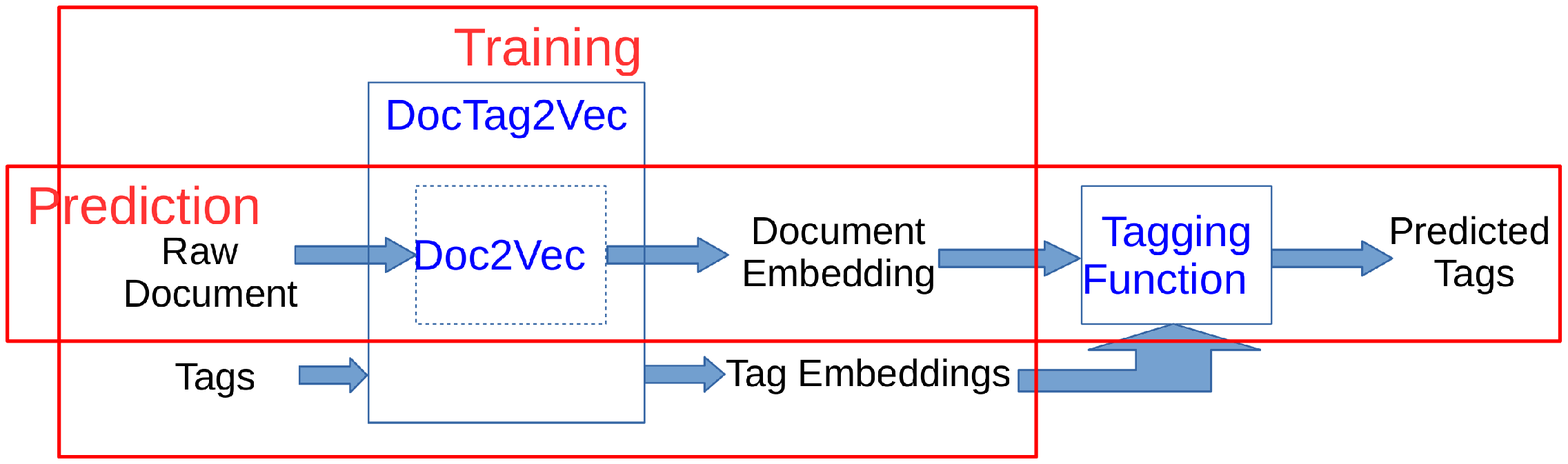}
        \caption{proposed method (DocTag2Vec)}
        \label{fig:sys2}
	\end{subfigure}%
    \vspace{-2mm}
\caption{Comparison of standard multi-label learning framework and the proposed method}
\vspace{-4mm}
\end{figure*}
From machine learning perspective, document tagging is by nature a \emph{multi-label learning} (MLL) problem, where the input space is certain feature space $\cal X$ of document and the output space is the power set $2^{\cal Y}$ of a finite set of tags $\cal Y$. Given training data $\cal Z \subset \cal X \times \text{$2$}^{\cal Y}$, we want to learn a function $f : \cal X \mapsto \text{$2$}^{\cal Y}$ that predicts tags for unseen documents. As shown in Figure \ref{fig:sys1}, during training a standard MLL algorithm (big blue box) one typically attempts to fit the prediction function (small blue box) into feature vectors of documents and the corresponding tags. Note that feature vectors are generated \emph{separately} before training, and tags for each document are encoded as a $|\cal Y|$-dimensional binary vector with one  representing the presence and zero otherwise. In prediction phase, the learned prediction function will output relevant tags for the input feature vector of an unseen document. Following such a paradigm, many generic algorithms have been developed for MLL \cite{webu11,prva14,bjkv15}. With a surge of text content created by users online, such as blog posts, Wikipedia entries, etc., the algorithms for document tagging has many challenges. Firstly, time sensitive news articles are generated on a daily basis, and it is important for an algorithm to assign tags before they loose freshness. Secondly, new tagged documents could be fed into the training system, thus incrementally adapting the system to new training data without re-training from scratch is also critical. Thirdly, we might face a very large set of candidate tags that can change dynamically, as new things are being invented.

In view of the aforementioned challenges, in this paper we propose a new and simple approach for document tagging: DocTag2Vec. Our approach is motivated by the line of works on learning distributed representation of words and documents, e.g., Word2Vec \cite{mscc13} and Doc2Vec (a.k.a. Paragraph Vector) \cite{lemi14}. Word2Vec and Doc2Vec aim at learning low-dimensional feature vectors (i.e., embeddings) for words and documents from large corpus in an \emph{unsupervised} manner, such that similarity between words (or documents) can be reflected by some distance metric on their embeddings. The general assumption behind Word2Vec and Doc2Vec is that more frequent co-occurrence of two words inside a small neighborhood of document should imply higher semantic similarity between them (see Section \ref{sec:w2v_d2v} for details). 
The DocTag2Vec extends this idea to document and tag by positing that document and its associated tags should share high semantic similarity, which allows us to learn the embeddings of tags along with documents (see Section \ref{sec:dt2v} for details). Our method has two striking differences compared with standard MLL frameworks: firstly, our method directly works with raw text and \emph{does not} need feature vectors extracted in advance. Secondly, our DocTag2Vec produces tag embeddings, which carry semantic information that are generally not available from standard MLL framework. During training, DocTag2Vec directly takes the raw documents and tags as input and learns their embeddings using \emph{stochastic gradient descent} (SGD). 
In terms of prediction, a new document will be first embedded using a Doc2Vec component inside the DocTag2Vec, and tags are then assigned by searching for the nearest tags embedded around the document. Overall the proposed approach has the following merits.
\begin{itemize}
\item The SGD training supports the incremental adjustment of DocTag2Vec to new data. 
\vspace{-2mm}
\item The prediction uses the simple $k$-nearest neighbor search among tags instead of documents, whose running time does not scale up as training data increase.
\vspace{-2mm}
\item Since our method represent each individual tag using its own embedding vector, it it easy to dynamically incorporate new tags. 
\vspace{-2mm}
\item The output tag embeddings can be used in other applications.
\end{itemize}

\textbf{Related Work}:
Multi-label learning has found several applications in social media and web, like sentiment and topic analysis~\cite{hpll13}, social text stream analysis~\cite{rplv14}, and online advertising~\cite{agpv13}.
MLL has also been applied to diverse Natural Language Processing (NLP) tasks. However to the best of our knowledge we are the first to propose embedding based MLL approach to a NLP task. 
MLL has been applied to Word Sense Disambiguation (WSD) problem for polysemic adjectives~\cite{boleda2007modelling}. \cite{huang2013sentiment} proposed a joint model to predict sentiment and topic for tweets and ~\cite{surdeanu2012multi} proposed a multi-instance MLL based approach for relation extraction with distant supervision.  

Recently learning embeddings of words and sentences from large unannotated corpus has gained immense popularity in many NLP tasks, such as Named Entity Recognition~\cite{passosetal2014,lampleetal2016,mahovy2016}, sentiment classification~\cite{socher2011semi,tang2014learning,dos2014deep} and summarization \cite{kaageback2014extractive,rush2015neural,li2015hierarchical}.
Also, vector space modeling has been applied to search re-targeting \cite{gdrn15} and query rewriting \cite{gdrs15}. 

Given many potential applications, document tagging has been a very active research area. In information retrieval, it is often coined as content-based \emph{tag recommendation} problem \cite{ccnh07}, for which numbers of approaches were proposed, such as \cite{herg08}, \cite{szlz08}, \cite{sozg08} and \cite{sozg11}. Personalized tag recommendation is also studied in the literature \cite{synm08,rbns09}. In machine learning community, a lot of general MLL algorithms have been developed, with application to document tagging, including compressed-sensing based approach \cite{hklz09}, WSABIE \cite{webu11}, ML-CSSP \cite{bikw13}, LEML \cite{yjkd14}, FastXML \cite{prva14}, SLEEC \cite{bjkv15} to name a few.

\textbf{Paper Organization}: The rest of the paper is organized as follows. In Section \ref{sec:approach}, we first give a brief review of Word2Vec and Doc2Vec models, and then present training and prediction step respectively for our proposed extension, DocTag2Vec. In Section \ref{sec:exp}, we demonstrate the effectiveness of our DocTag2Vec approach through experiments on several datasets. In the end, Section \ref{sec:conc} is dedicated to conclusions and future works.

\section{Proposed Approach}\label{sec:approach}
In this section, we present details of DocTag2Vec. For the ease of exposition, we first introduce some mathematical notations followed by a brief review for two widely-used embedding models: Word2Vec and Doc2Vec. 

\subsection{Notation}
We let $V$ be the size of vocabulary (i.e., set of unique words), $N$ be the number of documents in the training set, $M$ be the size of tag set, and $K$ be the dimension of the vector space of embedding. We denote the vocabulary as $\cW = \{w_1, \ldots, w_V \}$, set of documents as $\cD = \{d_1, \ldots, d_N\}$, and the set of tags as $\cT = \{t_1, \ldots, t_M\}$.  Each document $d \in \cD$ is basically a sequence of $n_d$ words represented by $(w^d_1, w^d_2, \ldots, w^d_{n_d})$, and is associated with $M_d$ tags $\cT_d = \{ t^d_1, \ldots, t^d_{M_d} \}$. Here the subscript $d$ of $n$ and $M$ suggests that the number of word and tag is different from document to document. For convenience, we use the shorthand $w^d_{i} : w^d_{j}$, $i \leq j$, to denote the subsequence of words $w^d_i, w^d_{i+1}, \ldots, w^d_{j-1}, w^d_j$ in document $d$. Correspondingly, we denote $\bW = [\w_1, \ldots, \w_V] \in \R^{K \times V}$ as the matrix for word embeddings, $\bD = [\d_1, \ldots, \d_N] \in \R^{K \times N}$ as the matrix for document embeddings, and $\bT = [\t_1, \ldots, \t_M] \in \R^{K \times M}$ as the matrix for tag embeddings. Sometimes we may use the symbol $d_i$ interchangeably with the embedding vector $\d_i$ to refer to the $i$-th document, and use $\d_d$ to denote the vector representation of document $d$. Similar conventions apply to word and tag embeddings. Besides we let $\sigma(\cdot)$ be the sigmoid function, i.e., $\sigma(a) = 1 / (1 + \exp(-a))$.

\subsection{Word2Vec and Doc2Vec}
\label{sec:w2v_d2v}
\begin{figure*}[hbt!]
	\centering
	\begin{subfigure}{0.29\textwidth}
		\centering
		\includegraphics[width=0.65\textwidth]{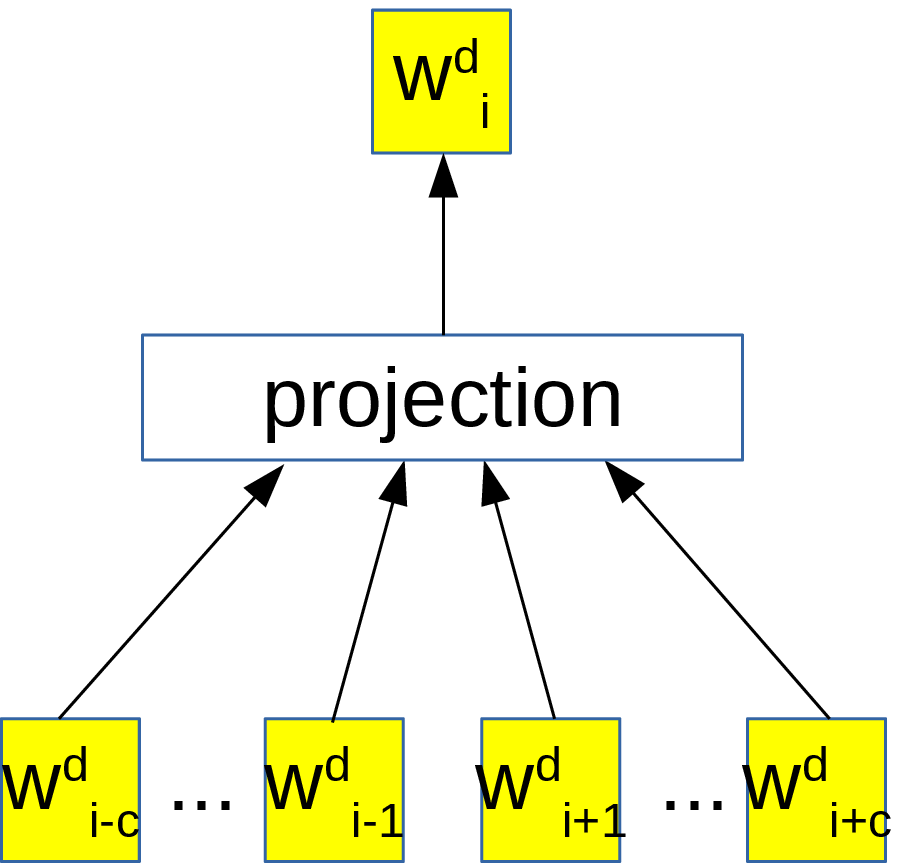}
        \caption{CBOW Word2Vec}
        \label{fig:word2vec}
	\end{subfigure}%
	\begin{subfigure}{0.29\textwidth}
		\centering
		\includegraphics[width=0.82\textwidth]{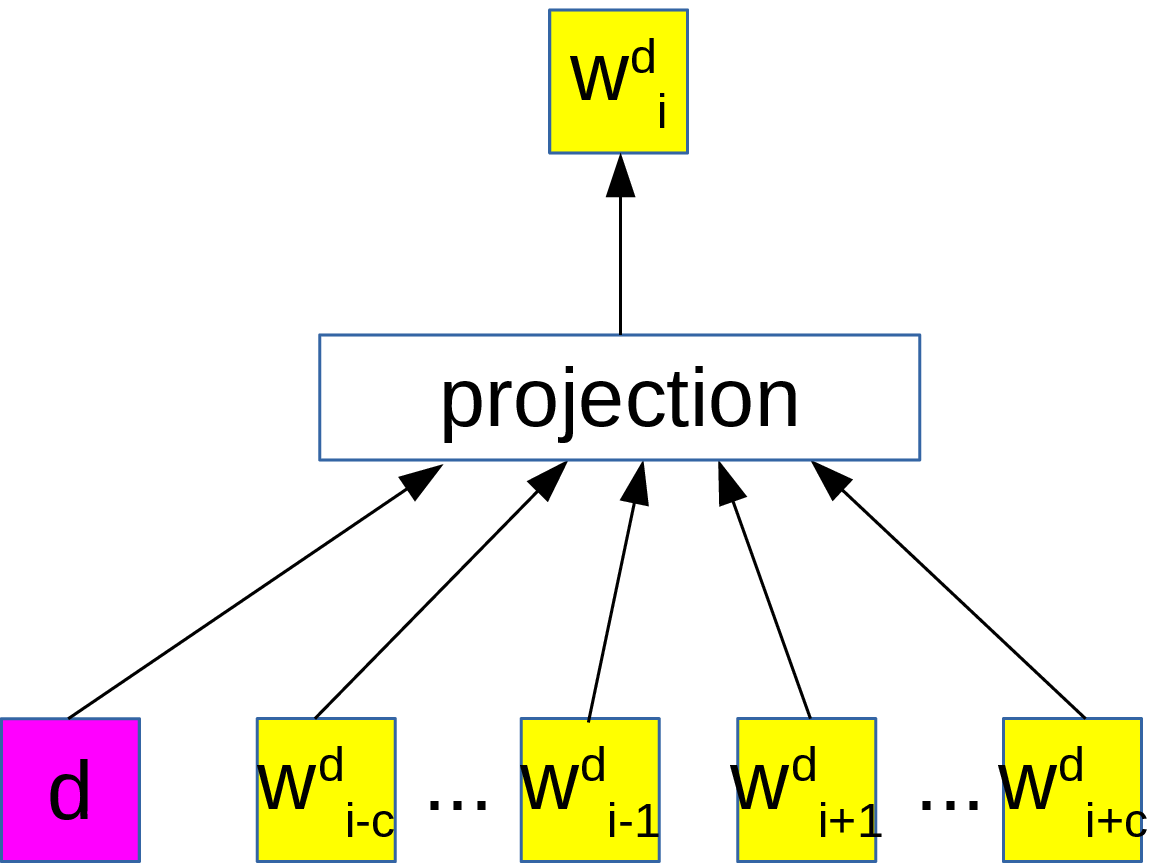}
        \caption{DM Doc2Vec}
        \label{fig:doc2vec}
	\end{subfigure}
	\begin{subfigure}{0.40\textwidth}
		\centering
		\includegraphics[width=0.9\textwidth]{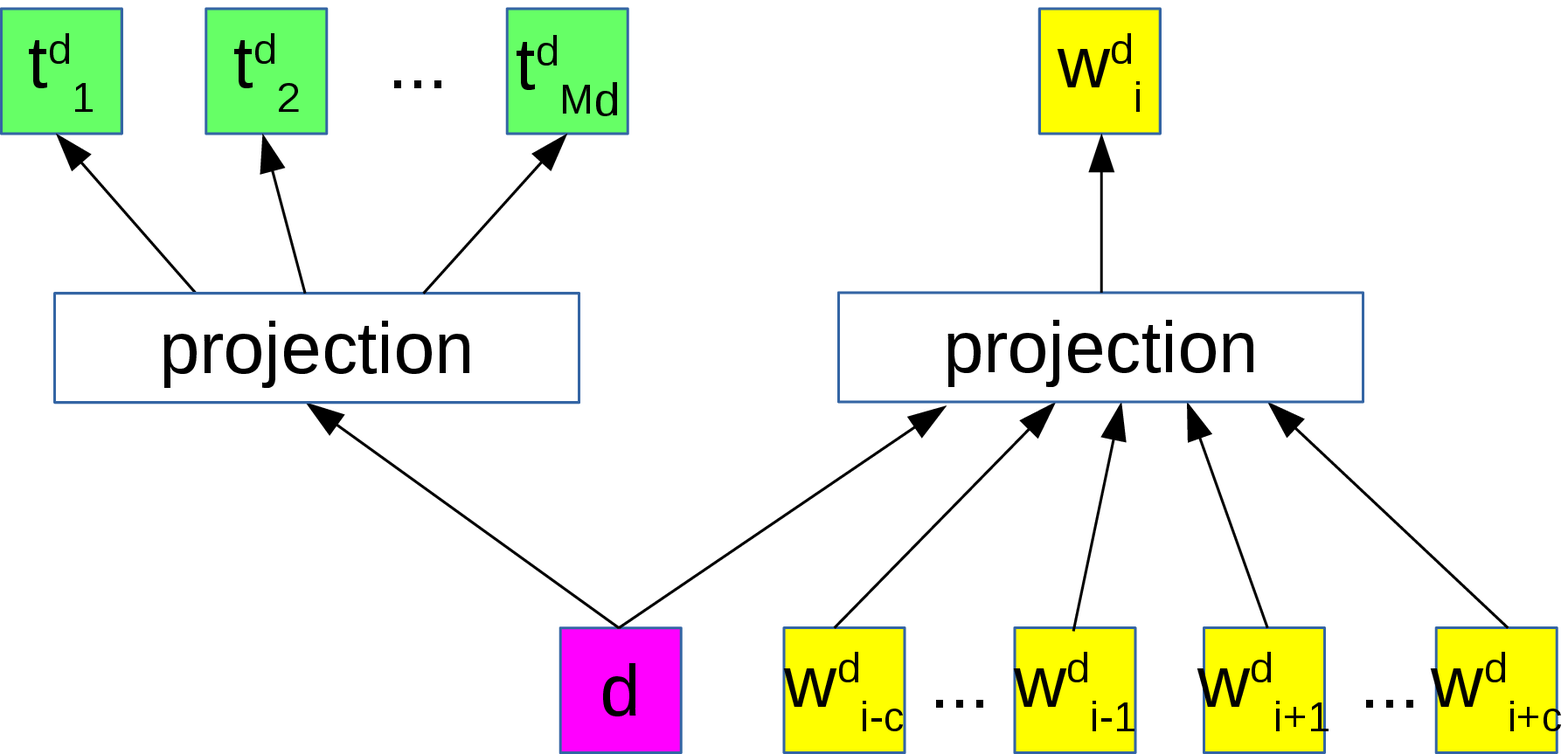}
        \caption{DocTag2Vec}
        \label{fig:doctag2vec}
	\end{subfigure}
\vspace{-2.5mm}
\caption{Model architectures of different embedding approaches}
\vspace{-2mm}
\end{figure*}

The proposed approach is inspired by the work of Word2Vec, an unsupervised model for learning embedding of words. Essentially, Word2Vec embeds all words in the training corpus into a low-dimensional vector space, so that the semantic similarities between words can be reflected by some distance metric (e.g., cosine distance) defined on their vector representations. The way to train Word2Vec model is to minimize the loss function associated with certain classifier with respect to both \emph{feature vectors} (i.e., word embeddings) and \emph{classifier parameters}, such that the nearby words are able to predict each other. For example, in \emph{continuous bag-of-word} (CBOW) framework, Word2Vec specifically minimizes the following average negative log probability 
\begin{align*}
\sum_{d \in \cD} \sum_{i=1}^{n_d} - \log p(w^d_{i} \ | \ w^d_{i-c}: w^d_{i-1}, w^d_{i+1}: w^d_{i+c}) ~,
\end{align*}
where $c$ is the size of context window inside which words are defined as ``nearby''. To ensure the conditional probability above is legitimate, one usually needs to evaluate a partition function, which may lead to a computationally prohibitive model when the vocabulary is large.
A popular choice to bypass such issue is to use hierarchical softmax (HS) \cite{mobe05}, which factorizes the conditional probability into products of some simple terms. The hierarchical softmax relies on the construction of a binary tree $\cB$ with $V$ leaf nodes, each of which corresponds to a particular word in the vocabulary $\cW$. HS is parameterized by a matrix $\bH \in \R^{K \times (V-1)}$, whose columns are respectively  mapped to a unique non-leaf node of $\cB$. Additionally, we define $\text{Path}(w) = \{ (i,j) \in \cB \ | \ \text{edge } (i,j) \text{ is on the path from root to word } w\}$. Then the negative log probability is given as
\begin{gather*}
\begin{split}
&\quad - \log p(w^d_{i} \ | \ w^d_{i-c}: w^d_{i-1}, w^d_{i+1}: w^d_{i+c}) \\
&= - \log \prod_{(u, v) \in \text{Path}(w^d_i)} \sigma \left( {\text{child}} (v)   \cdot \langle  \g^d_i, \h_{v} \rangle \right) \\
&= - \sum_{(u, v) \in \text{Path}(w^d_i)} \log \sigma \left( {\text{child}} (v)   \cdot \langle  \g^d_i, \h_{v} \rangle \right)  ~,
\end{split}
\\
\begin{split}
\g^d_i = \sum_{\substack{-c\leq j \leq c \\ j \neq 0}} \w^d_{i+j} ~,
\end{split}
\end{gather*}
where $\text{child}(u, v)$ is equal to $1$ if $v$ is the left child of $u$ and 0 otherwise. Figure \ref{fig:word2vec} shows the model architecture of CBOW Word2Vec. Basically $\g_i^d$ is the input feature for HS classifier corresponding to projection layer in Figure \ref{fig:word2vec} , which essentially summarizes the feature vectors of context words surrounding $w^d_i$, and other options like averaging of $\w^d_{i+j}$ can also be applied. This Word2Vec model can be directly extended to Distributed memory (DM) Doc2Vec model by conditioning the probability of $w^d_i$ on $d$ as well as $w^d_{i-c}, \ldots, w^d_{i+c}$, which yields
\begin{gather}
\begin{split}
\label{eq:doc_hs}
&\quad -\log p(w^d_{i} \ | \ w^d_{i-c} : w^d_{i-1}, w^d_{i+1} : w^d_{i+c}, d) \\
&= - \sum_{(u, v) \in \text{Path}(w^d_i)} \log \sigma \left( {\text{child}} (v)   \cdot \langle  \tilde{\g}^d_i, \h_{v} \rangle \right) ~,
\end{split}
\\
\begin{split}
\tilde{\g}^d_i = \d_d + \sum_{\substack{-c\leq j \leq c \\ j \neq 0}} \w^d_{i+j} ~.
\end{split}
\end{gather}
The architecture of DM Doc2Vec model is illustrated in Figure \ref{fig:doc2vec}. Instead of optimizing some rigorously defined probability function, both Word2Vec and Doc2Vec can be trained using other objectives, e.g., negative sampling (NEG) \cite{mscc13}.

\subsection{Training for DocTag2Vec}
\label{sec:dt2v}
Our approach, DocTag2Vec, extends the DM Doc2Vec model by adding another component for learning tag embeddings. In addition to predicting target word $w^d_i$ using context $w^d_{i-c}, \ldots, w^d_{i+c}$, as shown in Figure \ref{fig:doctag2vec}, DocTag2Vec also uses the document embedding to predict each associated tag, with hope that they could be closely embedded. The joint objective is given by
\begin{align*}
\sum_{d \in \cD} \sum_{i=1}^{n_d} \Big ( &- \log p(w^d_{i} \ | \ w^d_{i-c} : w^d_{i-1}, w^d_{i+1} : w^d_{i+c}, d)  \\
&- \alpha \sum_{t \in \cT_d} \log p(t \ | \ d) \Big ) ~,
\end{align*}
where $\alpha$ is a tuning parameter. As discussed for Word2Vec, the problem of evaluating costly partition function is also faced by the newly introduced probability $p(t | d)$. Different from the conditional probability of $w^d_i$, the probability $p(t | d)$ cannot be modeled using hierarchical softmax, as the columns of parameter matrix do not have one-to-one correspondence to tags (remember that we need to obtain a vector representation for each tag). Motivated by the idea of negative sampling used in Word2Vec, we come up with the following objective for learning tag embedding rather than stick to a proper probability function,
\begin{align}
\begin{split}
\label{eq:tag_ns}
- \sum_{t \in \cT_d}&\log \sigma \left( \langle \d_t, \t_t \rangle \right) +  r \cdot \E_{\t \sim p} \left[\log \sigma\left(- \langle \d_t, \t \rangle \right) \right] ~,
\end{split}
\end{align}
where $p$ is a discrete distribution over all tag embeddings $\{\t_1, \ldots, \t_M\}$ and $r$ is a integer-valued hyperparameter. The goal of such objective is to differentiate the tag $t$ from the draws according to $p$, which is chosen as \emph{uniform distribution} for simplicity in our practice. Now the final loss function for DocTag2Vec is the combination of \eqref{eq:doc_hs} and \eqref{eq:tag_ns},
\begin{align}
\begin{split}
\label{eq:overall_obj}
&\ell\left(\bW, \bD, \bT, \bH \right) = \sum_{d \in \cD} \sum_{i=1}^{n_d} \\ 
&\Big ( \underbrace{-\sum_{\substack{(u, v) \in \text{Path}(w^d_i)}} \log \sigma \big( {\text{child}} (v)   \cdot \langle  \tilde{\g}^d_i, \h_{v} \rangle \big)}_{\text{DM Doc2Vec with hierarchical softmax}} \ - \\
& \underbrace{\alpha \sum_{t \in \cT_d}  \log \sigma \left( \langle \d_t, \t_t \rangle \right) + r \cdot \E \left[\log \sigma\left(- \langle \d_t, \t \rangle \right) \right]}_{\text{tag embedding with negative sampling}} \Big )
\end{split}
\end{align}
We minimize $\ell\left(\bW, \bD, \bT, \bH \right)$ using stochastic gradient descent (SGD). 
To avoid exact calculation of the expectation in negative sampling, at each iteration we sample $r$ i.i.d. instances of $\t$ from distribution $p$, denoted by $\{\t^1_{p}, \t^2_{p}, \ldots, \t^r_{p} \}$,  to stochastically approximate the expectation, i.e., $\sum_{j=1}^r \log \sigma(-\langle \d_t, \t_p^j \rangle) \approx r \cdot \E \left[\log \sigma\left(- \langle \d_t, \t \rangle \right) \right]$. 

\subsection{Prediction for DocTag2Vec}
Unlike Word2Vec and Doc2Vec, which only target on learning high-quality embeddings of words and documents, DocTag2Vec needs to make predictions of relevant tags for new documents. To this end, we first embed the new document via the Doc2Vec component within DocTag2Vec and then perform $k$-nearest neighbor ($k$-NN) search among tags. To be specific, given a new document $d$, we first optimize the objective \eqref{eq:doc_hs} with respect to $\d_d$ by fixing $\bW$ and $\bH$. Note that this is the standard inference step for new document embedding in Doc2Vec. Once $\d_d$ is obtained, we search for the $k$-nearest tags to it based on cosine similarity. Hence the prediction function is given as
\begin{align}
\begin{split}
f_k(\d_d) = \big \{ i \left. \right\vert  &\text{$u_i$ is in the largest $k$ entries} \\ 
&\text{of $\u=\overline{\bT}^T \overline{\d}_d$} \ \big \} ~,
\end{split}
\end{align}
where $\overline{\bT}$ is column-normalized version of $\bT$. To boost the prediction performance of DocTag2Vec, we apply the \emph{bootstrap aggregation} (a.k.a. bagging) technique to DocTag2Vec. Essentially we train $b$ DocTag2Vec learners using different randomly sampled subset of training data, resulting in $b$ different tag predictors $f_{k'}^1(\cdot), \ldots, f_{k'}^b(\cdot)$ along with their tag embedding matrices $\bT^1, \ldots, \bT^b$. In general, the number of nearest neighbors $k'$ for individual learner can be different from $k$. In the end, we combine the predictions from different models by selecting from $\bigcup_{j=1}^b f_{k'}^j(\d_d)$ the $k$ tags with the largest aggregated similarities with $\d_d$,
\begin{align*}
\begin{split}
f^{bag}_k(\d_d) = \Big \{ i  \left. \right\vert &\text{$u_i$ is in the largest $k$ entries of $\u$}, \\ 
\text{where } u_i = & \sum_{j=1}^b \I\{i \in f^j_{k'}(\d_d)\} \cdot \langle \overline{\t}_i^{j}, \overline{\d}_d \rangle \ \Big \} ~.
\end{split}
\end{align*}

\begin{table*}[hbt!]
\small
\centering
\begin{tabular}{|c|c|c|c|c|}
\hline 
Dataset & \#training point & \#testing point & \#unique tags & Avg \#tags per document  \\ 
\hline
Wiki10 & 14095 & 6600 & 31177 & 17.27 \\
\hline
WikiNER & 89521  & 10000 & 67179 & 22.96 \\
\hline 
Relevance Modeling (Chinese) & 4505 &  500 & 391 & 1.02 \\ 
\hline 
Relevance Modeling (Korean) & 1292 & 500 & 261 & 1.07 \\
\hline
NCT (all) & 40305 & 9546 & 883 & 1.88 \\
\hline
NCT (general) & 39401 & 9389 &  294 & 1.76 \\
\hline
NCT (specific) & 17278 & 4509 &  412  & 1.41 \\
\hline
\end{tabular} 
\vspace{-1mm}
\caption{Statistics of datasets}
\label{tab:datastat}
\vspace{-2mm}
\end{table*}

\begin{figure*}[t]
	\centering      
	\begin{subfigure}{0.24\textwidth}
		\centering
		\includegraphics[width=0.95\textwidth]{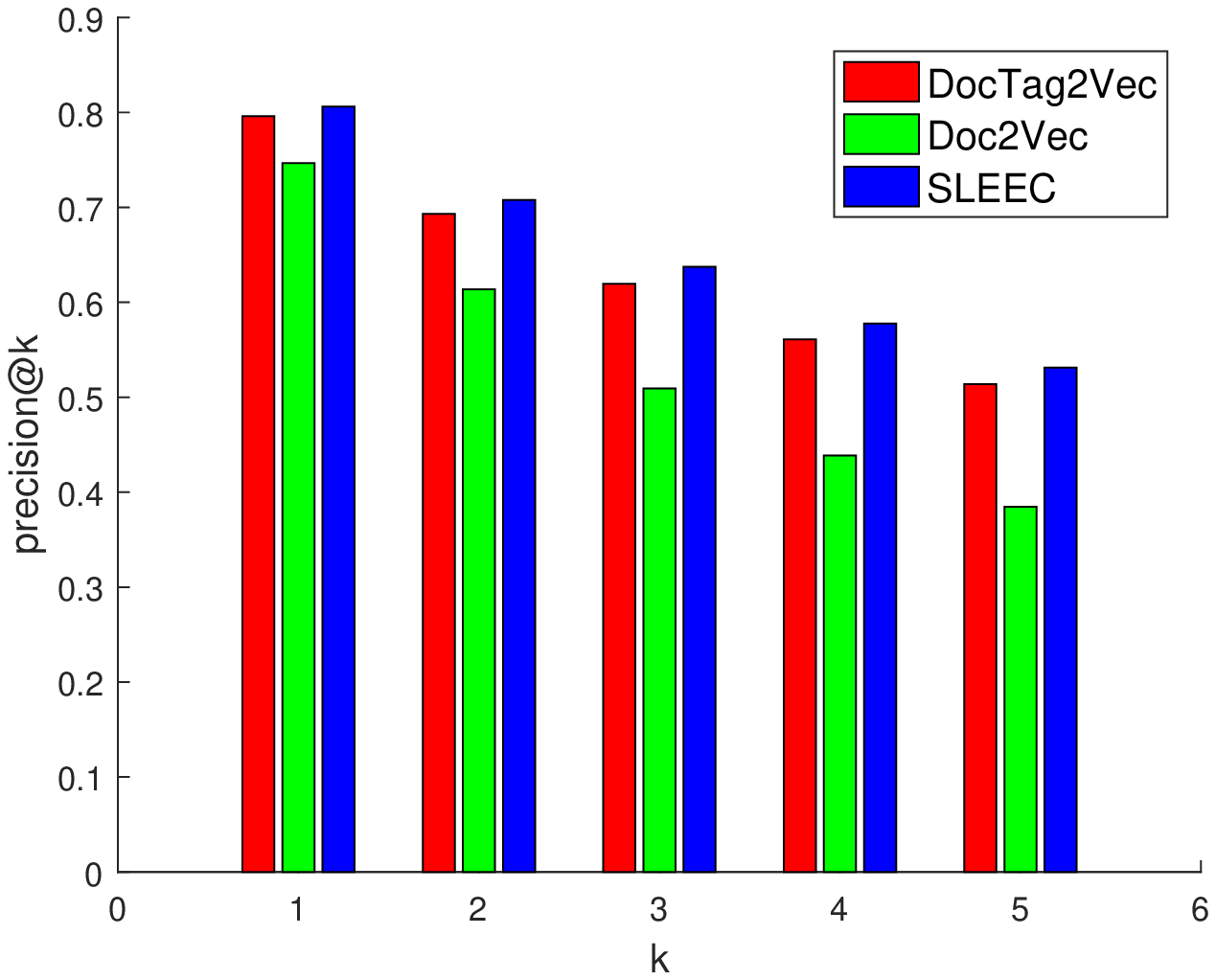}
        \caption{Wiki10}
        \label{fig:wiki10}
	\end{subfigure}
	\begin{subfigure}{0.24\textwidth}
		\centering
		\includegraphics[width=0.95\textwidth]{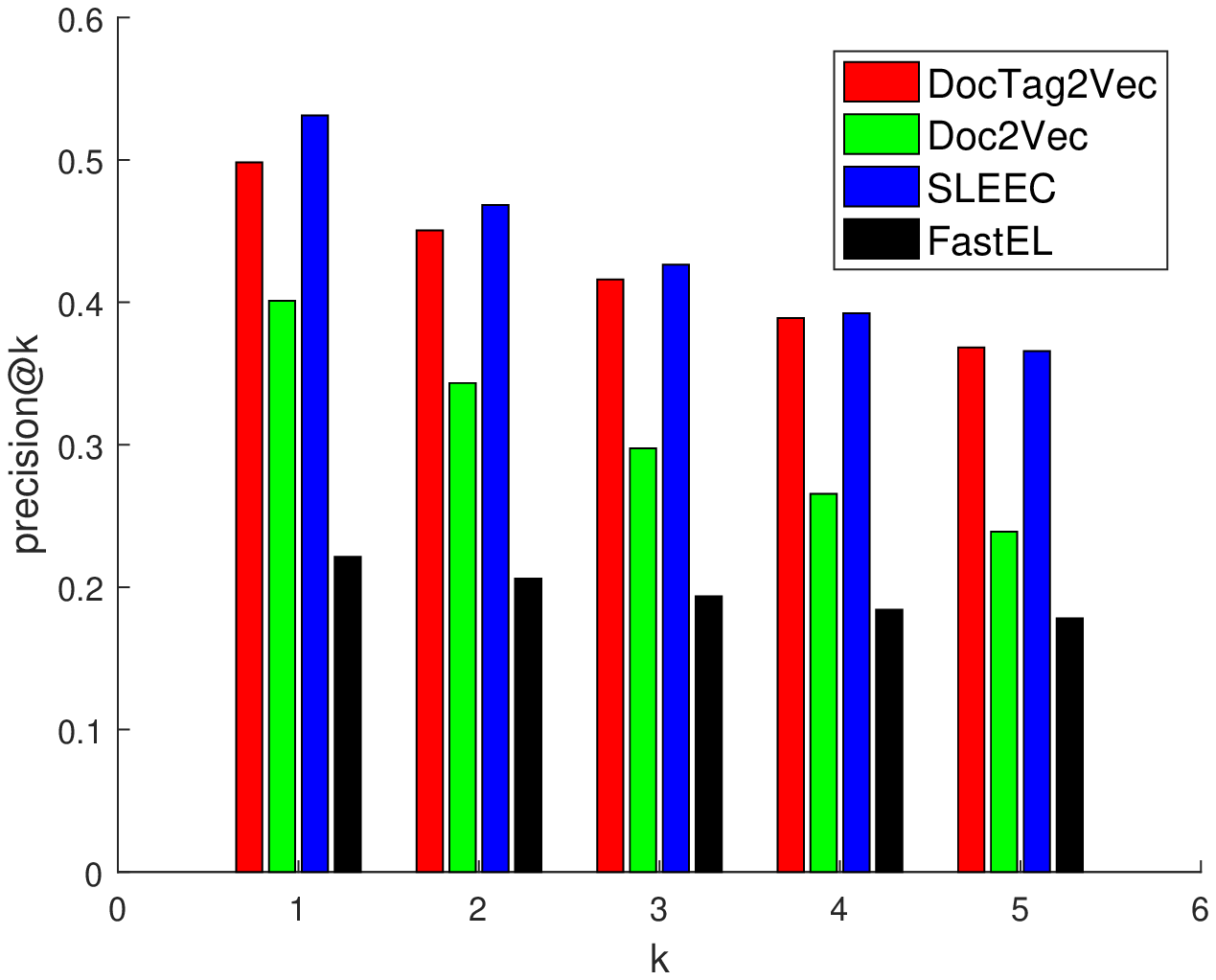}
        \caption{WikiNER}
        \label{fig:wikiner}
	\end{subfigure}
	\begin{subfigure}{0.24\textwidth}
		\centering
		\includegraphics[width=0.95\textwidth]{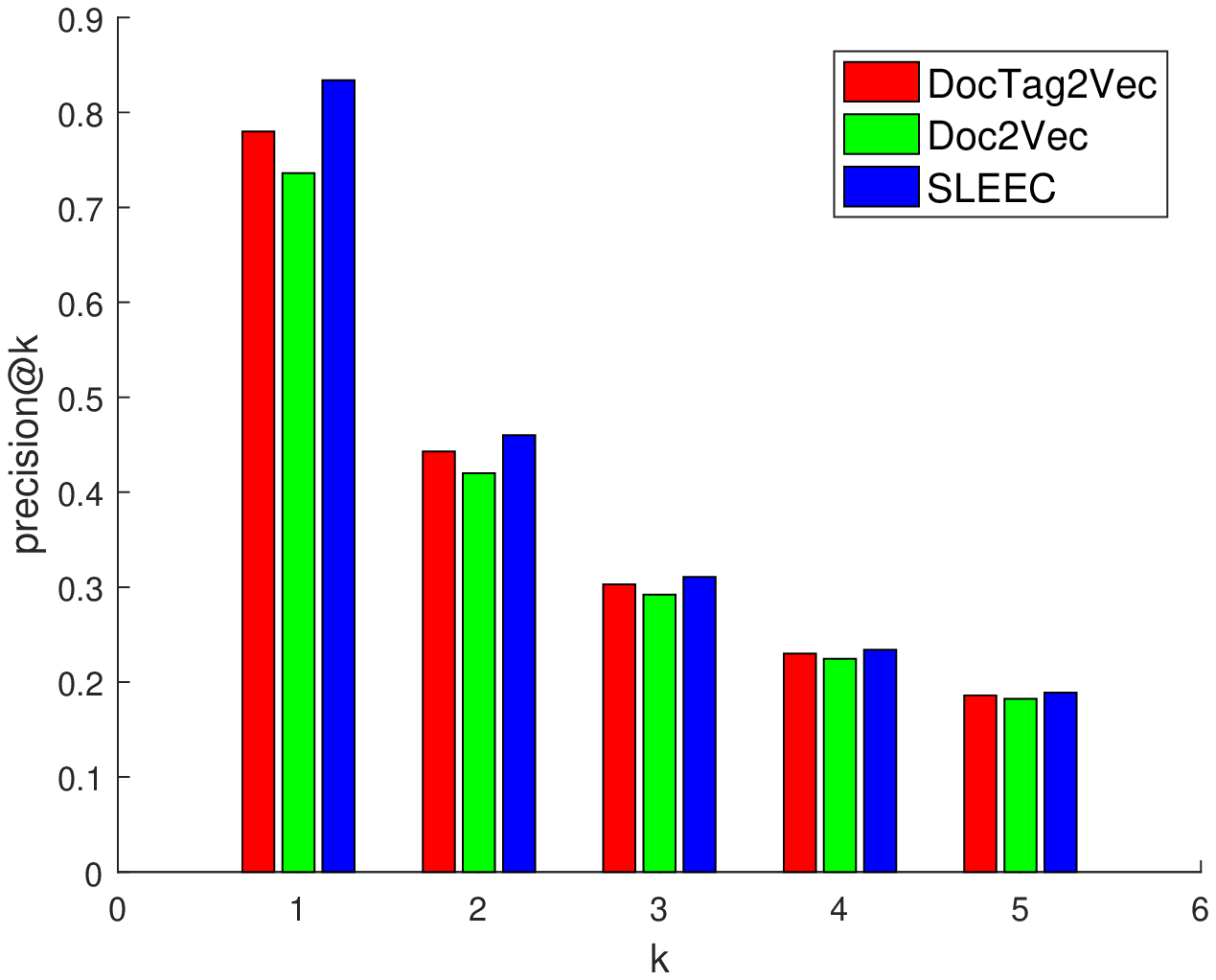}
        \caption{RM (Chinese)}
        \label{fig:CN}
	\end{subfigure}
	\begin{subfigure}{0.24\textwidth}
		\centering
		\includegraphics[width=0.95\textwidth]{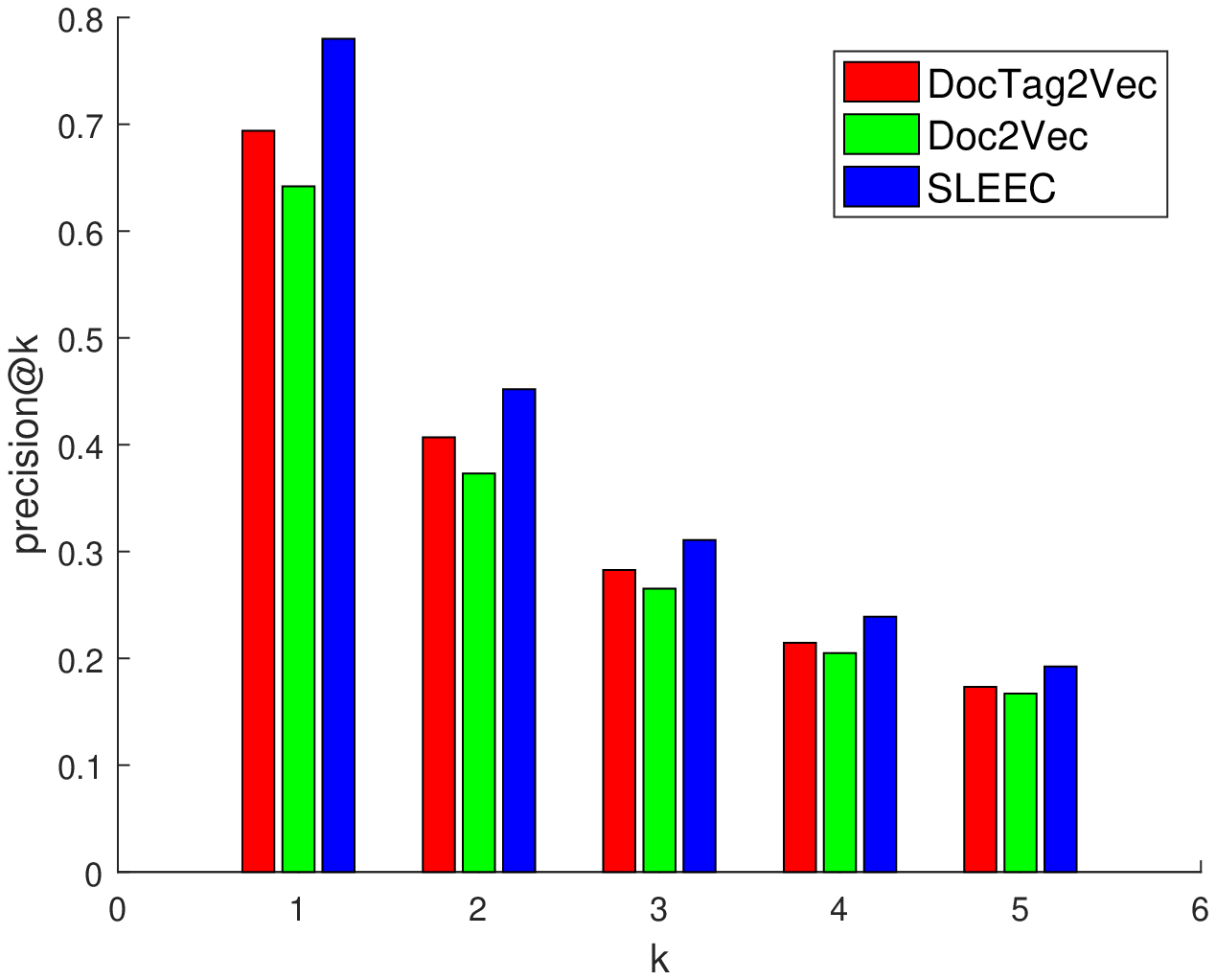}
        \caption{RM (Korean)}
        \label{fig:KO}
	\end{subfigure}
\vspace{-2mm}
\caption{Precision on Wiki10, WikiNER and Relevance Modeling dataset}
\label{fig:rest_dataset}
\vspace{-2mm}
\end{figure*}

\begin{figure*}[t]
	\centering      
	\begin{subfigure}{0.32\textwidth}
		\centering
		\includegraphics[width=0.95\textwidth]{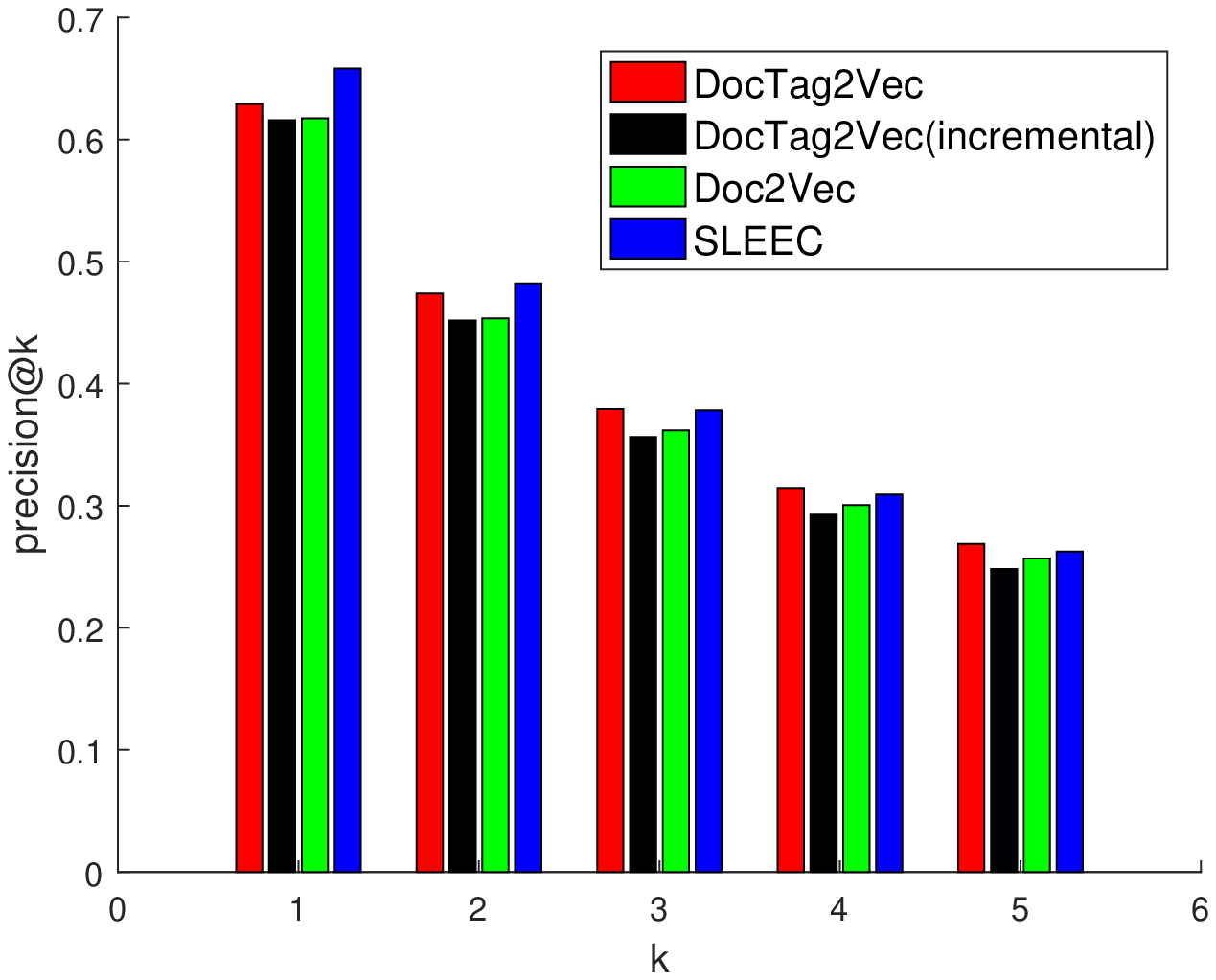}
        \caption{All tags}
        \label{fig:yct_all}
	\end{subfigure}
	\begin{subfigure}{0.32\textwidth}
		\centering
		\includegraphics[width=0.95\textwidth]{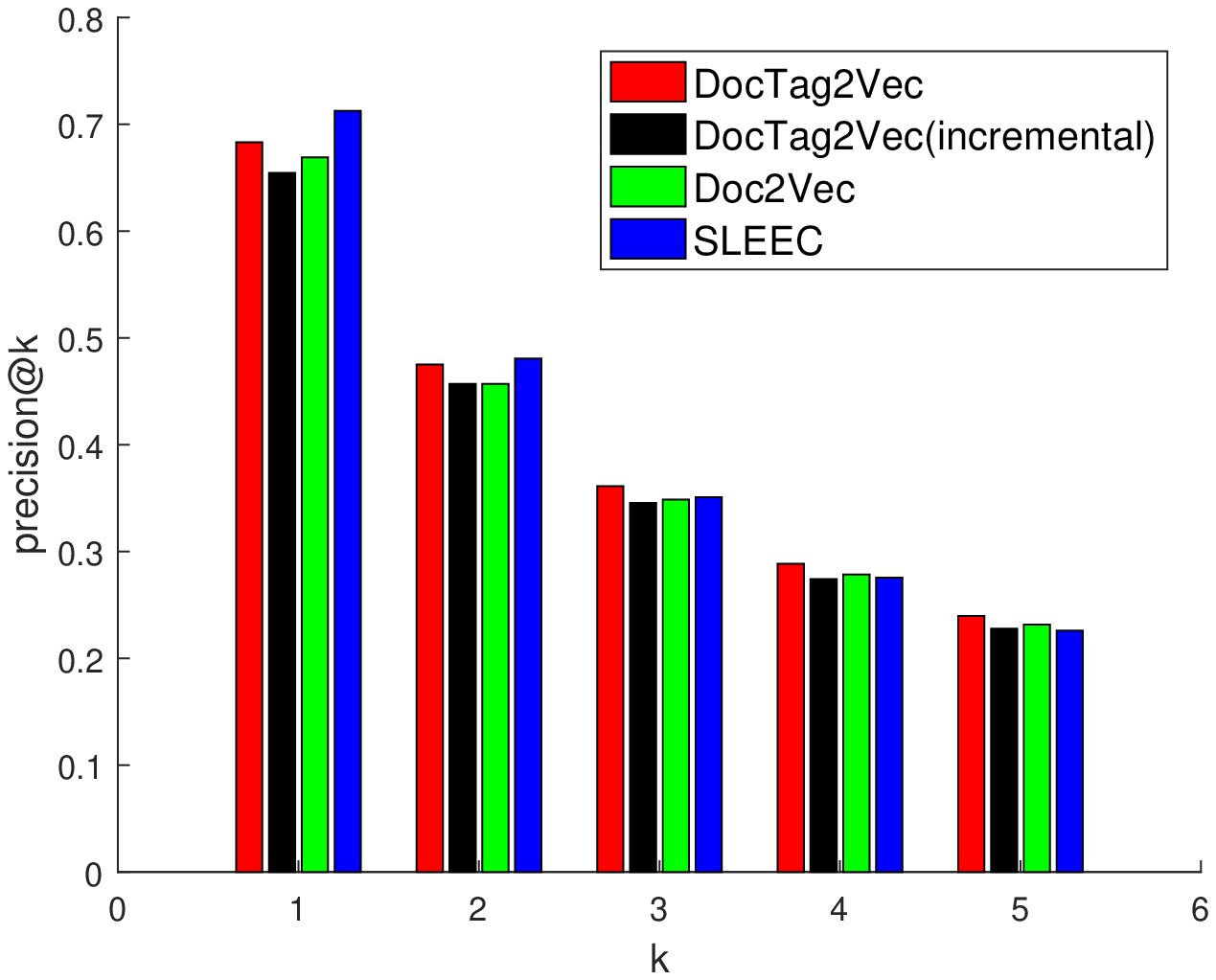}
        \caption{Specific tags}
        \label{fig:yct_specific}
	\end{subfigure}
	\begin{subfigure}{0.32\textwidth}
		\centering
		\includegraphics[width=0.95\textwidth]{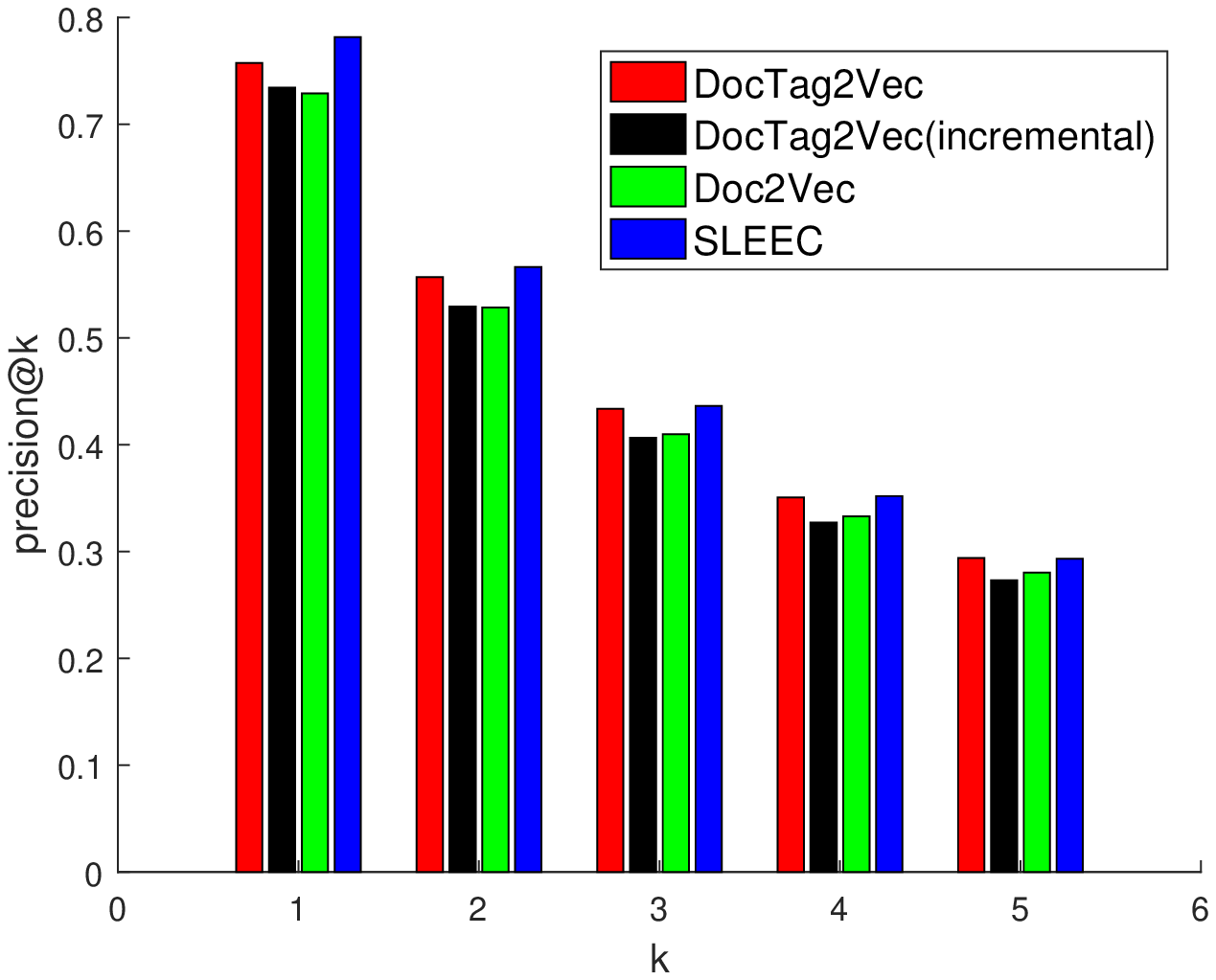}
        \caption{General tags}
        \label{fig:yct_general}
	\end{subfigure}
\vspace{-2mm}
\caption{Precision on News Content Taxonomy dataset}
\label{fig:yct}
\vspace{-2mm}
\end{figure*}

\begin{table*}
\small
\centering
\begin{tabular}{|c|c|c|c|c|}
  \hline
  \ & DocTag2Vec & DocTag2Vec (incremental) & Doc2Vec & SLEEC \\
  \hline
  NCT (all tags) & 0.6702 & 0.6173 & 0.6389 & 0.6524 \\
  \hline
  NCT (specific tags) & 0.8111 & 0.7678 & 0.7810 & 0.7624 \\
  \hline
  NCT (general tags) & 0.7911 & 0.7328 & 0.7521 & 0.7810 \\
  \hline
\end{tabular}
\caption{Overall Recall on News Content Taxonomy dataset}
\label{tab:yct_recall}
\end{table*}

\begin{figure*}[hbt!]
	\centering
	\includegraphics[width=0.95\textwidth]{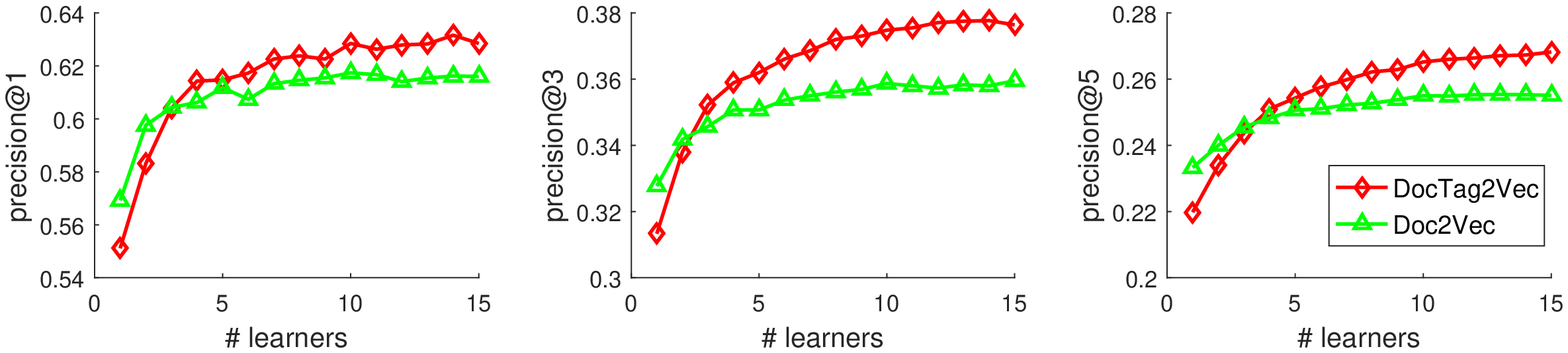}
    \vspace{-2mm}
	\caption{Precision vs. number of learners on NCT dataset}
	\label{fig:yct_learner}
\vspace{-2mm}
\end{figure*}

\begin{figure*}[hbt!]
	\centering
	\includegraphics[width=0.95\textwidth]{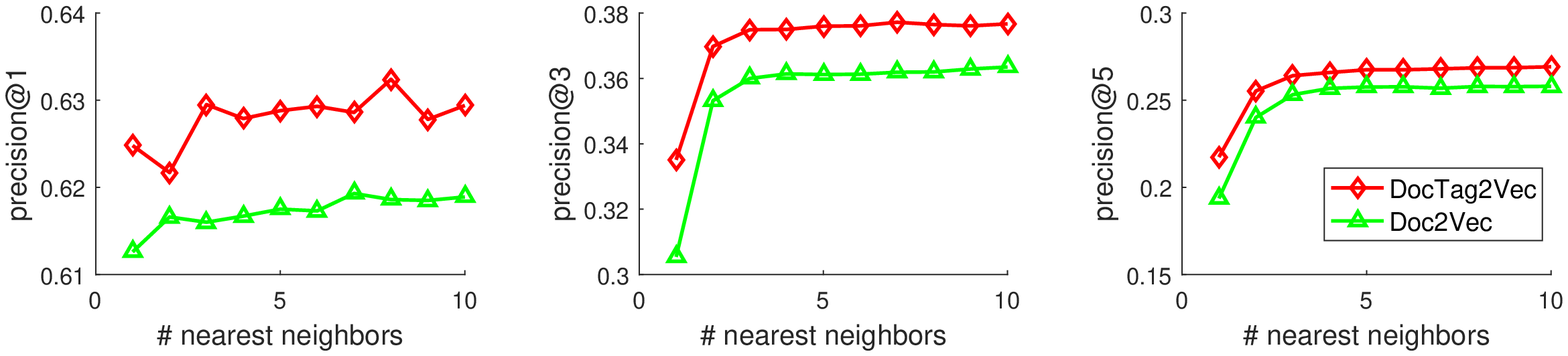}
    \vspace{-2mm}
	\caption{Precision vs. number of nearest neighbors on NCT dataset}
	 \label{fig:yct_knn}
\vspace{-1mm}
\end{figure*}

\section{Experiments}
\label{sec:exp}
\subsection{Datasets}
In this subsection, we briefly describe the datasets included in our experiments. It is worth noting that DocTag2Vec method needs raw texts as input instead of extracted features. Therefore many benchmark datasets for evaluating multi-label learning algorithms are not suitable for our setting. For the experiment, we primarily focus on the diversity of the source of tags, which capture different aspects of documents. The statistics of all datasets are provided in Table
\ref{tab:datastat}. \vspace{+2mm} \\
\textbf{Public datasets}:
\begin{itemize}
\item
\textbf{Wiki10}:
The wiki10 dataset contains a subset of English Wikipedia documents, which are tagged collaboratively by users from the social bookmarking site \emph{Delicious}\footnote{https://del.icio.us/}. We remove the two uninformative tags, ``wikipedia'' and ``wiki'', from the collected data. 
\item
\textbf{WikiNER}: 
WikiNER has a larger set of English Wikipedia documents. The tags for each document are the named entities inside it, which is detected automatically by some named entity recognition (NER) algorithm. 
\end{itemize} 
\textbf{Proprietary datasets}
\begin{itemize}
\item
\textbf{Relevance Modeling (RM)}: 
The RM dataset consists of two sets of financial news article in Chinese and Korean respectively. Each article is tagged with related ticker symbols of companies given by editorial judgement.
\item
\textbf{News Content Taxonomy (NCT)}: NCT dataset is a collection of news articles annotated by editors with topical tags from a taxonomy tree. The closer the tag is to the root, the more general the topic is. For such tags with hierarchical structure, we also evaluate our method separately for tags of general topics (depth=2) and specific topics (depth=3). 
\end{itemize}

\subsection{Baselines and Hyperparameter Setting}
The baselines include one of the state-of-the-art multi-label learning algorithms called SLEEC \cite{bjkv15}, a variant of DM Doc2Vec, and an unsupervised entity linking system, FastEL~\cite{blanco2015fast}, 
which is specific to WikiNER dataset. 
SLEEC is based on non-linear dimensionality reduction of binary tag vectors, and use a sophisticated objective function to learn the prediction function. For comparison, we use the TF-IDF representation of document as the input feature vector for SLEEC, as it yields better result than embedding based features like Doc2Vec feature. To extend DM Doc2Vec for tagging purpose, basically we replace the document $d$ shown in Figure \ref{fig:doc2vec} with tags $t^d_1, \ldots, t^d_{M_d}$, and train the Doc2Vec to obtain the tag embeddings. During testing, we perform the same steps as DocTag2Vec to predict the tags, i.e., inferring the embedding of test document followed by $k$-NN search.  FastEL is unsupervised appproach for entity linking of web-search queries that  walks over a sequence of words in query and aims to maximize the likelihood of linking the text span to an entity in Wikipedia. FastEL model calculates the conditional probabilities of an entity given every substring of the input document, however avoid computing entit to entity joint dependencies, thus making the process efficient. We built FastEL model using query logs that spanned 12 months and Wikipedia anchor text extracted from Wikipedia dumps dated November 2015. We choose an entity linker baseline because it is a simple way of detecting topics/entities that are semantically associated with a document. 

Regarding hyperparameter setting, both SLEEC and DocTag2Vec aggregate multiple learners to enhance the prediction accuracy, and we set the number of learners to be 15. For SLEEC, we tune the rest of hyperparameters using grid search. For SLEEC and DocTag2Vec, we set the number of epochs for SGD to be 20 and the window size $c$ to be 8. To train each individual learner, we randomly sample $50\%$ training data. In terms of the nearest neighbor search, we set $k' = 10$ for Wiki10 and WikiNER while keeping $k'=5$ for others.  For the rest of hyperparameters, we also apply grid search to find the best ones. For DocTag2Vec, we additionally need to set the number of negative tags $r$ and the weight $\alpha$ in 
\eqref{eq:overall_obj}. Typically $r$ ranges from 1 to 5, and $r=1$ gives the best performance on RM and NCT datasets. Empirically good choice for $\alpha$ is between 0.5 and 5.  For FastEL, we consider a sliding window of size $5$ over the raw-text (no punctuations) of document to generate entity candidates. We limit the number of candidates per document to 50. 

\begin{table*}[htb!]
\scriptsize
\centering
\begin{tabular}{|m{8.0cm}|m{2.5cm}|m{2.5cm}|m{1.0cm}|}
\hline 
\multirow{3}{*}{ \quad \quad \quad \quad \quad \quad \quad \quad \quad \quad \quad \quad \quad News excerpt} & \multirow{3}{*}{\ \ \ Editorial tags}  & \multicolumn{2}{c|}{Prediction (top 3)} \\ 
\cline{3-4}
\ & \ & \quad \ Predicted tags & similarity \\
\hline
\vspace{-0.65cm}
\multirow{3}{8cm}{The world is definitely getting warmer, according to the U.S. National Atmospheric and Oceanic Administration. For its annual "State of the Climate" report, NOAA for the first time gathered data on 37 climate indicators, such as air and sea temperatures, sea level, humidity, and snow cover in one place, and found that, taken together, the measurements show an "unmistakable upward trend" in temperature. Three hundred scientists analyzed the information and concluded it's "undeniable" that the planet has warmed since 1980, with the last decade taking the record for hottest ever recorded. }  & \multirow{3}{2.5cm}{\textit{/Nature \& Environment/ Natural Phenomena}} & \textit{/Nature \& Environment/ Environment/Climate Change} & 1.99 \\
\cline{3-4}
 &  &  \textit{/Science/ Meteorology} \newline & 0.64 \\
\cline{3-4}
 &  & \textit{/Nature \& Environment/Natural Phenomena/Weather} & 0.57 \\
\hline 
\vspace{-0.5cm}
\multirow{3}{8cm}{Business software maker Epicor Software Corp. said Thursday that its second-quarter loss narrowed as revenue climbed. For the April-June quarter, Epicor's loss totaled \$1 million, or 2 cents per share, compared with a loss of \$6.7 million, or 11 cents per share, in the year-ago quarter. When excluding one-time items, Epicor earned 13 cents per share, which is what analysts polled by Thomson Reuters expected. Revenue rose 9 percent to \$109.2 million, beating analyst estimates for \$105.2 million.} & \vspace{-0.5cm} \multirow{3}{2.5cm}{\textit{/Business/Sectors \& Industries/ Information Technology/Internet Software \& Services \newline \newline /Finance/Investment \& Company Information}}  & \textit{ /Finance/Investment \& Company Information/ Company Earnings} & 2.25 \\
\cline{3-4}
 &  &  \textit{/Finance/Investment \& Company Information}   & 1.23 \\
\cline{3-4}
 &  &  \textit{/Finance/Investment \& Company Information/ Stocks \& Offerings} & 0.32 \\
\hline
\vspace{-0.3cm}
\multirow{3}{8.0cm}{TicketLiquidator, the leading provider of the world's most extensive ticket inventory for hard-to-find, low priced tickets, today announced that tickets are available for the Orlando Magic vs. Cleveland Cavaliers game on Wednesday, November 11th at Orlando's Amway Arena. The much-anticipated matchup features LeBron James, who is now in the final year of his contract with the Cavaliers. } & \multirow{3}{2.5cm}{\textit{/Sports \& Recreation/ Baseball}} & \textit{/Sports \& Recreation/Basketball } & 4.07 \\
\cline{3-4}
 &  &  \textit{/Arts \& Entertainment/Events/Tickets} & 3.42 \\
\cline{3-4}
 &  &  \textit{/Sports \& Recreation/Baseball} & 0.96 \\
\hline
\end{tabular}
\vspace{-2mm}
\caption{Examples of Better Prediction over Editorial Judgement}
\label{tab:good_mistake}
\vspace{-4mm}
\end{table*}

\subsection{Results}
	
We use $precision@k$ as the evaluation metric for the performance.
Figure \ref{fig:rest_dataset} shows the precision plot of different approaches against choices of $k$ on Wiki10, WikiNER and RM dataset. On Wiki10, we see that the precision of our DocTag2Vec is close to the one delivered by SLEEC, while Doc2Vec performs much worse. We observe similar result on WikiNER except for the precision@1, but our precision catches up as $k$ increases. For RM dataset, SLEEC outperforms our approach, and we conjecture that such gap is due to the small size of training data, from which DocTag2Vec is not able to learn good embeddings. It is to be noted that SLEEC requires proper features as input and does not work directly with raw documents; while DocTag2Vec learns vector representation of documents that are not only useful for multilabel learning but also as features for other tasks like sentiment analysis, hate speech detection, and content based recommendation. We have demonstrated improvements in all the above mentioned use-cases of DocTag2Vec vectors but the discussion on those is out of the scope of this paper. 

For NCT dataset, we also train the DocTag2Vec incrementally, i.e., each time we only feed 100 documents to DocTag2Vec and let it run SGD, and we keep doing so until all training samples are presented. As shown in Figure \ref{fig:yct}, our DocTag2Vec outperform Doc2Vec baseline, and delivers competitive or even better precision in comparision with SLEEC. Also, the incremental training does not sacrifice too much precision, which makes DocTag2Vec even appealing. The overall recall of DocTag2Vec is also slightly better than SLEEC, as shown in Table \ref{tab:yct_recall}. Figure \ref{fig:yct_learner} and \ref{fig:yct_knn} include the precision plot against the number of learners $b$ and the number of nearest neighbors $k'$ for individual learner, respectively. It is not difficult to see that after $b=10$, adding more learners does not give significant improvement on precision. For nearest neighbor search, $k'=5$ would suffice.

\subsection{Case Study for NCT dataset}

For NCT dataset, when we examine the prediction for individual articles, it turns out surprisingly that there are a significant number of cases where DocTag2Vec outputs better tags than those by editorial judgement. Among all these cases, we include a few in Table \ref{tab:good_mistake} showing the superiority of the tags given by DocTag2Vec sometimes. For the first article, we can see that the three predicted tags are all related to the topic, especially the one with highest similarity, \textit{/Nature \& Environment/ Environment/Climate Change}, seems more pertinent compared with the editor's. Similarly, we predict \textit{/Finance/Investment \& Company Information/Company Earnings} as the most relevant topic for the second article, which is more precise than its parent \textit{/Finance/Investment \& Company Information}. Besides our approach can even find the wrong tags assigned by the editor. The last piece of news is apparently about NBA, which should have the tag  \textit{/Sports \& Recreation/Basketball} as predicted, while the editor annotates them with the incorrect one, \textit{/Sports \& Recreation/Baseball}. On the other hand, by looking at the similarity scores associated with the predicted tags, we can see that higher score in general implies higher aboutness, which can also be used as a quantification of prediction confidence.

\section{Conclusions and Future Work}\label{sec:conc}
In this paper, we present a simple method for document tagging based on the popular distributional representation learning models, Word2Vec and Doc2Vec. Compared with classical multi-label learning 
methods, our approach provides several benefits, such as allowing incremental update of model, handling the dynamical change of tag set, as well as producing feature representation for tags. 
The document tagging can benefit a number of applications on social media. If the text content over web is correctly tagged, articles or blog posts can be pushed to the right users who are likely to be interested. And such good personalization will potentially improve the users engagement.
In future, we consider extending our approach in a few directions.  
Given that tagged documents are often costly to obtain, it would be interesting to extend our approach to a semi-supervised setting, where we can incorporate large amounts of unannotated documents to enhance our model. On the other hand, with the recent progress in graph embedding for social network \cite{yacs16,grle16}, we may be able to improve the tag embedding by exploiting the representation of users and interactions between tags and users on social networks.

\section*{Acknowledgements}
The authors would like to thank the anonymous reviewers for their encouraging and thoughtful comments.

\bibliography{eacl2017}
\bibliographystyle{acl_natbib}

\end{document}